\pdfoutput=1
\documentclass[10pt,twocolumn,letterpaper]{article}

\usepackage[pagenumbers]{cvpr}

\usepackage{caption}
\usepackage{times}
\usepackage{latexsym}
\usepackage{amsmath,amssymb}
\usepackage{array,booktabs,multirow,colortbl}
\usepackage{float}
\usepackage{placeins}
\usepackage[T1]{fontenc}
\usepackage{inconsolata}
\usepackage[most]{tcolorbox}
\lstdefinestyle{promptjson}{%
  basicstyle=\footnotesize\ttfamily,
  breaklines=true,
  breakatwhitespace=true,
  columns=fullflexible,
  keepspaces=true,
  showstringspaces=false,
  upquote=true,
  aboveskip=4pt,
  belowskip=4pt
}
\newtcolorbox{annotationprompt}{%
  enhanced,
  colback=black!1,
  colframe=black!25,
  colbacktitle=black!1,
  coltitle=black,
  title={Pairwise CoT Annotation Prompt},
  fonttitle=\small\bfseries,
  fontupper=\fontsize{10}{10.5}\selectfont,
  sharp corners,
  boxrule=0.35pt,
  titlerule=0.35pt,
  left=2mm,
  right=2mm,
  top=1mm,
  bottom=1mm,
  toptitle=1mm,
  bottomtitle=1mm,
  before skip=5pt,
  after skip=5pt
}

\definecolor{cvprblue}{rgb}{0.21,0.49,0.74}
\usepackage[pagebackref,breaklinks,colorlinks,allcolors=cvprblue]{hyperref}

\title{VersusQ: Pairwise Margin Reasoning \\for Generalizable Video Quality Assessment}

\author{
Shibei Meng\textsuperscript{1}\quad
Binxin Yang\textsuperscript{2}\quad
Yuan Liu\textsuperscript{3}\quad
Jiexuan Zhang\textsuperscript{4}\\[2pt]
Zhengyao Lv\textsuperscript{5}\quad
Hubery Yin\textsuperscript{2}\quad
Qiang Xu\textsuperscript{1}\\[6pt]
{\normalsize
\textsuperscript{1}The Chinese University of Hong Kong\quad
\textsuperscript{2}WeChat AI\quad
\textsuperscript{3}Beijing Normal University}\\
{\normalsize
\textsuperscript{4}Peking University\quad
\textsuperscript{5}The University of Hong Kong}
}

\begin{document}
\maketitle
\begin{abstract}
Large Multimodal Models (LMMs) have shown promise for video quality assessment, but most methods still predict an absolute score for each video. Such pointwise supervision often mixes perceptual quality with dataset-specific calibration, including annotation protocols, rating habits, and score distributions. As a result, the learned scoring rule may work well within a benchmark but transfer poorly across unseen domains. We argue that relative comparisons alleviate the absolute-scale calibration bias by focusing purely on perceptual differences rather than dataset-specific rating habits. Consequently, we propose \textbf{VersusQ}, a pairwise margin reasoning framework driven entirely by direct comparisons. Specifically, VersusQ performs LMM-based comparison between two videos, reasons about their visual and temporal quality differences, and predicts a signed continuous margin that captures both the preferred choice and the degree of difference. 
Furthermore, to align interpretable comparison rationales with fine-grained numerical differences, we introduce Margin-Coupled GRPO, which jointly optimizes rollout-based relational reasoning and continuous margin regression. Extensive experiments on multiple public VQA benchmarks demonstrate that VersusQ achieves state-of-the-art performance, strong cross-domain generalization, and reliable fine-grained ranking under heterogeneous evaluation scenarios.
\end{abstract}
\section{Introduction}
\label{sec:intro}

\begin{figure}[t]
  \centering
  \includegraphics[width=\linewidth]{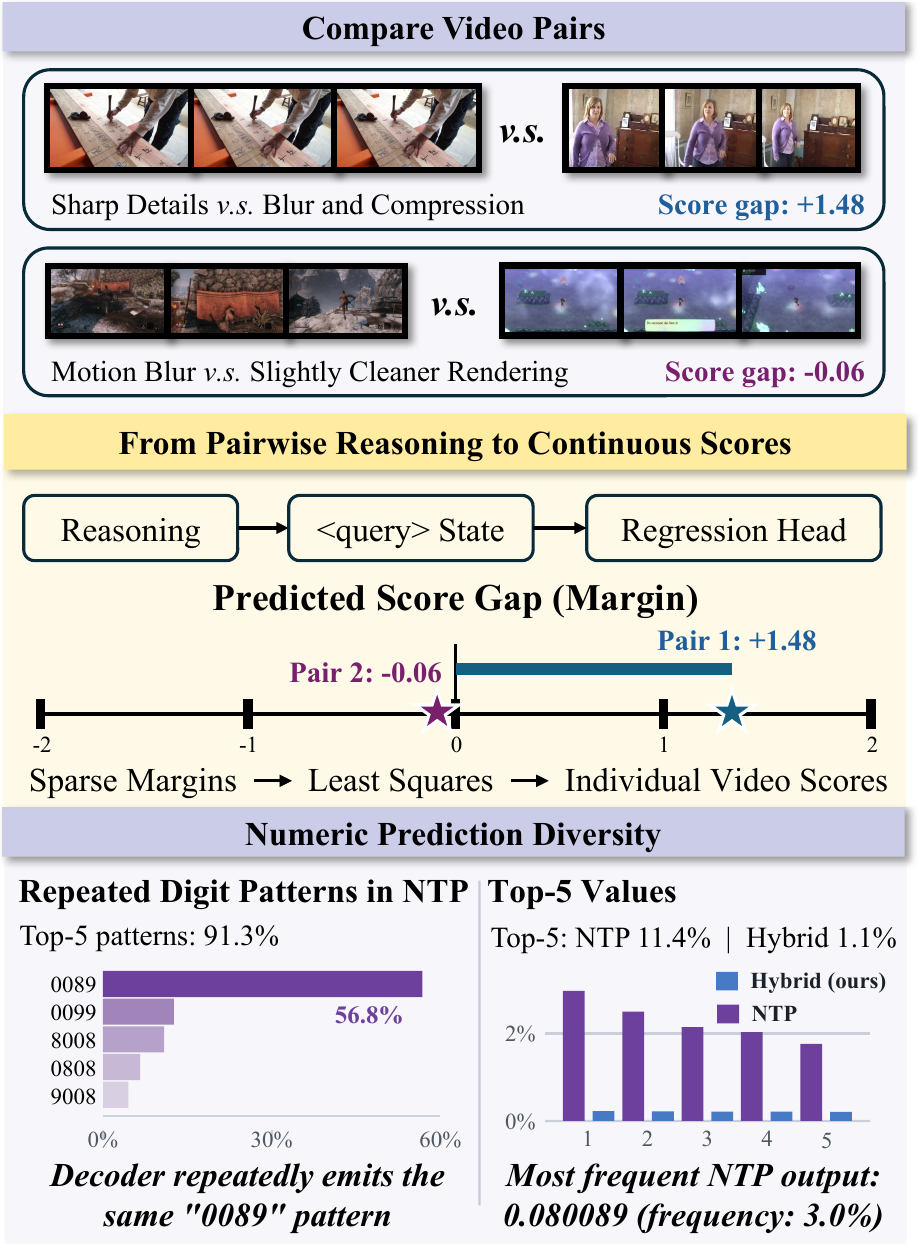}
  \caption{\textbf{Pairwise margins and numeric diversity.} Top: Video pairs and score gaps (left minus right). Middle: Reasoning and score reconstruction. Bottom: Repeated digit patterns in next-token prediction (NTP) and top-5 value frequencies for NTP versus hybrid regression.}
  \label{fig:ntp-vs-hybrid-collapse}
  \vspace{-5pt}
\end{figure}
Image Quality Assessment (IQA) and Video Quality Assessment (VQA) aim to predict how humans perceive the visual quality of a given image or video, serving as a foundational building block of modern multimedia systems~\citep{min2024perceptual, zhai2023blind}. As video sources grow increasingly diverse in recording environments, compression methods, and content types, a practical VQA model must produce assessments that remain faithful across unseen domains and diverse degradations. Recent progress in large multimodal models (LMMs) has motivated a series of LMM-based I/VQA methods~\citep{qalign,deqascore,qinsight,visualqualityr1,vqathinker}. Despite these advances, a central challenge remains: \textit{generalization}. Most approaches evaluate each video independently. Such pointwise scoring forces models to learn an absolute quality scale from heterogeneous datasets with varying rating protocols and score distributions, making global comparison across benchmarks inherently difficult.

\textit{Pairwise comparison} is a natural and important formulation for quality assessment. It has a long history in ranking and preference modeling~\citep{bradley1952rank,burges2005learning,liu2009learning}, and has also been explored in IQA/VQA~\citep{rankiqa,zhu2024compare2score}. The appeal is clear: comparing two videos removes the need to commit to a dataset-specific bias, encourages the model to focus on \textit{relative perceptual evidence}, and better matches how humans judge \textit{subtle quality differences}. Figure~\ref{fig:ntp-vs-hybrid-collapse} illustrates how signed margins from video pairs yield a continuous score scale and contrasts the numeric predictions of next-token prediction (NTP) and hybrid regression.

Despite its clear conceptual appeal, fully transitioning from isolated pairwise experiments to a generalizable continuous VQA leaderboard still gives rise to two central challenges:
\textit{\textbf{(1) How to Design an Efficient Pairwise Comparison Framework?}} 
Going from shattered pairwise comparisons back to a continuous leaderboard is expensive. Exhaustively enumerating all N-choose-2 pairs is prohibitive, and rating schemes like Elo~\citep{elo1978rating} still require a large number of random comparisons to converge.
\textit{\textbf{(2) How to Construct an Effective Reasoning Paradigm?}} 
A model that only outputs a binary preference explains neither why nor by how much one video wins, making it prone to spurious shortcuts. This calls for interpretable perceptual reasoning chains that can produce a globally calibrated quality scale.

For the first challenge, a binary win or loss signal is too weak to rebuild a continuous scale efficiently. We therefore assign each comparison a \textbf{Predictive Margin}, a signed real value capturing both the direction and magnitude of the quality gap, so a sparse connected set of pairs is sufficient. At inference, a sparse connected set of these margins is consolidated via a closed-form least-squares solution~\citep{jiang2011statistical} to efficiently recover the global score for every video at once.

Furthermore, recognizing that discrete language tokens are ill-suited for fine-grained numerical scoring, we introduce \textbf{Margin-Coupled GRPO (MC-GRPO)}. This strategy attaches a lightweight regression head to the hidden states after the generated rationale. By jointly accumulating the policy gradient for reasoning and the continuous margin-regression gradient within the same optimization step, MC-GRPO ensures that fine-grained numerical evaluations are inherently backed by explicit perceptual rationales.

Our contributions are summarized as follows:
\begin{itemize}
    \item We propose a generalizable \textbf{pairwise margin ranking framework} built on multimodal relational reasoning, reducing the reliance on domain-specific absolute scores and focusing the model on learning transferable relative comparison patterns.
    \item Moving beyond conventional reasoning-only GRPO paradigms, we introduce \textbf{MC-GRPO}, which jointly optimizes the shared reasoning backbone and a margin regression head, ensuring fine-grained numerical predictions are grounded in interpretable relational rationales.
    \item Extensive experiments demonstrate that our approach achieves state-of-the-art performance across multiple mainstream benchmarks, delivering strong cross-domain generalization and reliable fine-grained ranking under heterogeneous evaluation scenarios.
\end{itemize}

\section{Related Work}
\label{sec:related_work}

\paragraph{Image and Video Quality Assessment.}
Image Quality Assessment (IQA) and Video Quality Assessment (VQA) have progressed from classical metrics such as PSNR, SSIM~\citep{wang2004image}, MOVIE~\citep{seshadrinathan2010motion}, VMAF~\citep{li2016toward}, and ST-RRED~\citep{soundararajan2013video} to CNN- and transformer-based models~\citep{kang2014convolutional, zhang2018blind, vsfa, fastvqa, dover, kvq} trained on large-scale perceptual annotations.
Large multimodal models (LMMs)~\citep{liu2023llava, chen2024internvl, bai2023qwenvl} further enable language-conditioned quality assessment: Q-Align~\citep{qalign} maps discrete textual levels to continuous scores, DeQA-Score~\citep{deqascore} softens level labels to reduce discretization loss, and VQA$^2$~\citep{vqa2scorer} builds instruction data for joint quality scoring and understanding.
Recent methods also elicit explicit quality reasoning with GRPO, including Q-Insight~\citep{qinsight}, VisualQuality-R1~\citep{visualqualityr1}, and VQAThinker~\citep{vqathinker}, while RALI~\citep{rali} uses reasoning traces as intermediate representations for prediction.
Nevertheless, most methods still train and score videos independently, limiting fine-grained adaptation to new domains. Compare2Score~\citep{zhu2024compare2score} mitigates this issue with pairwise comparison, but its direct five-level quantization limits granularity and interpretability.

\vspace{-10pt}
\paragraph{Reinforcement Learning for Large Multimodal Models.}
Reinforcement learning from human feedback (RLHF) aligns language models with human preferences by fitting reward models on pairwise comparisons under the Bradley--Terry formulation~\citep{christiano2017deep, ouyang2022training, bradley1952rank}.
Group Relative Policy Optimization (GRPO)~\citep{shao2024deepseekmath, deepseekr1} removes the separate critic model and estimates advantages from group-level rewards, making reasoning-oriented fine-tuning more efficient.
GRPO and related RL pipelines have been adopted for multimodal reasoning: Video-R1~\citep{videor1} applies rule-based RL to video understanding, OneThinker~\citep{onethinker} unifies image and video reasoning across diverse tasks, and general-purpose models such as InternVL3.5~\citep{internvl35} and Qwen3-VL~\citep{qwen3vl} incorporate multi-stage RL into training.
Beyond general understanding, Visual-RFT~\citep{visualrft} uses verifiable visual rewards for detection and classification, R1-Onevision~\citep{r1onevision} bridges visual perception and deep reasoning through SFT and RL, and Seg-Zero~\citep{segzero} trains reasoning-guided segmentation with GRPO alone.
Visual generation studies also address exploration~\citep{hu2026embeddingperturbed}, preference mode collapse~\citep{chen2026taming}, and spatial credit assignment~\citep{hu2026blameerrantwords}.
For quality assessment, however, existing RL pipelines for LMMs mainly optimize discrete textual outputs, leaving regression-oriented quality assessment as token-level score generation.
\section{Methodology}
\label{sec:method}

\begin{figure*}[t]
  \centering
  \includegraphics[width=\textwidth]{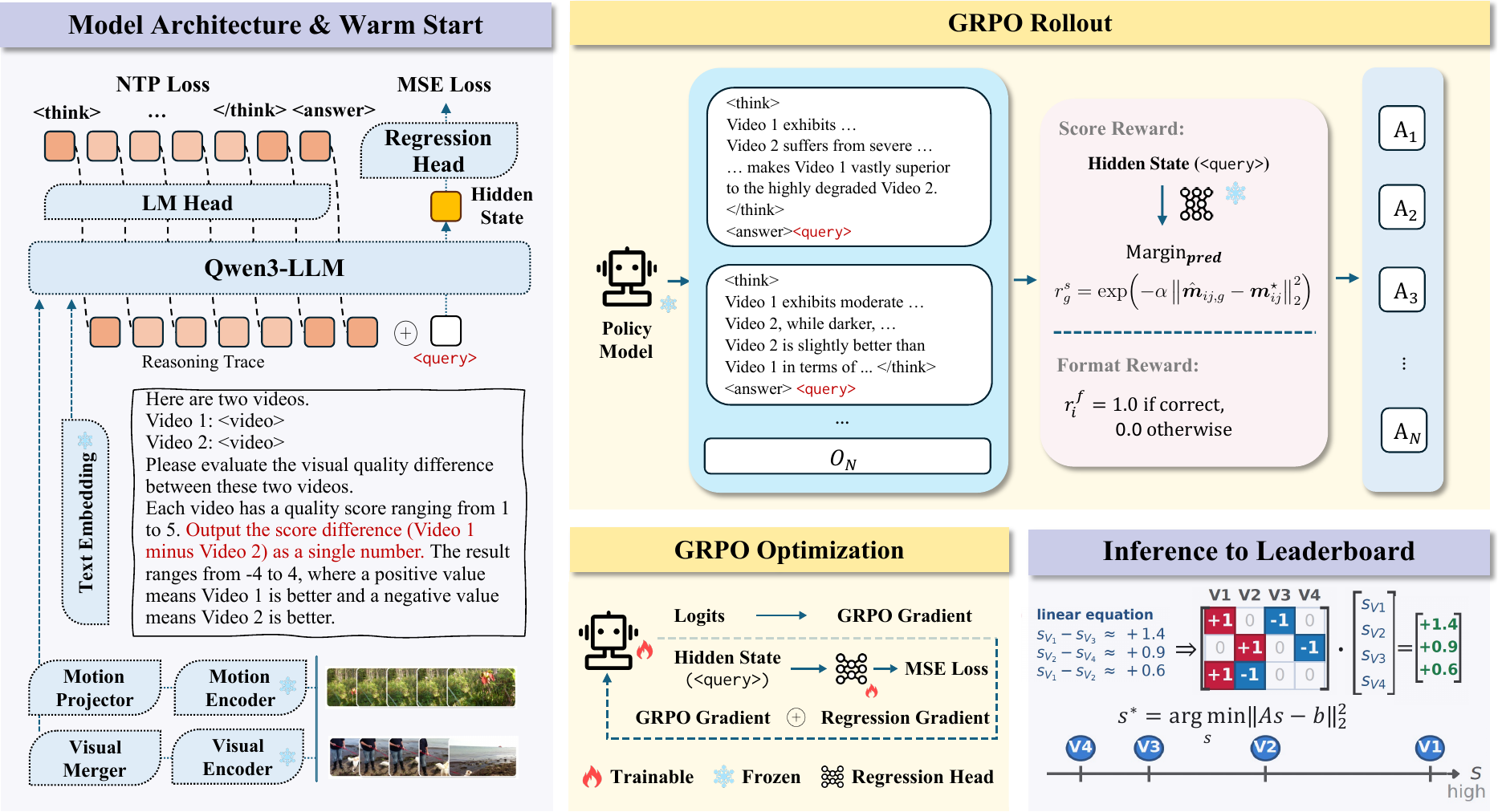}
  \caption{\textbf{Overview of the proposed VersusQ pipeline.} VersusQ first warm-starts a motion-aware VLM using rationale supervision and MSE margin regression derived from the designated \texttt{<query>} hidden state. Subsequently, MC-GRPO refines the reasoning process via score and format rewards, while employing an auxiliary regression gradient for calibration. At inference, predicted sparse margins are aggregated to construct an anchor-free leaderboard via zero-mean least squares.}
  \label{fig:framework}
\end{figure*}
\label{sec:method:overview}

\subsection{Overview}

To establish a generalizable framework for pairwise video quality assessment, \textbf{VersusQ} (Figure~\ref{fig:framework}) organizes the pipeline from supervision to optimization. 
\textbf{(1)}~\emph{Pairwise ranking with predictive margins} defines the signed quality-difference target $\boldsymbol{m}_{ij}^{\star}$ and reconstructs an anchor-free global leaderboard from predicted sparse margins via least-squares aggregation. 
\textbf{(2)}~\emph{Supervision for quality differences} specifies how the VLM processes each video pair: a motion-aware pathway injects temporal features into the language model, and a regression head predicts the margin under a hybrid reasoning-and-regression objective. 
\textbf{(3)}~\emph{Reinforced reasoning for quality assessment} improves comparison rationales with GRPO while preserving margin calibration through an auxiliary regression gradient. 
This organization separates the ranking formulation, the supervised quality-difference learner, and the reinforcement stage that aligns free-form reasoning with quantitative assessment. 

\subsection{Pairwise Ranking with Predictive Margins}
\label{sec:method:pairwise}

\label{sec:method:pairwise:margin}
\label{sec:method:pairwise:lsq}

Existing pairwise IQA methods such as Compare2Score~\citep{zhu2024compare2score} already move beyond binary win/lose labels by quantizing pairwise quality differences into discrete comparative levels and recovering scores with anchor-based probabilistic aggregation. However, this discretization still coarsens fine-grained differences and relies on predefined anchors. To retain continuous pairwise supervision, we define the target margin $\boldsymbol{m}_{ij}^{\star}\in\mathbb{R}$ as the signed difference between observed Mean Opinion Scores (MOS):
\begin{equation}
\label{eq:margin}
\boldsymbol{m}_{ij}^{\star} \;=\; s_i^{\mathrm{MOS}} \;-\; s_j^{\mathrm{MOS}}\,,
\end{equation}
where $s_i^{\mathrm{MOS}}$ and $s_j^{\mathrm{MOS}}$ are the ground-truth MOS annotations of two videos. The pairwise module predicts $\hat{\boldsymbol{m}}_{ij}$ toward this target, capturing both preference direction and quality-gap magnitude.

At inference time, the model predicts sparse relative margins $\{\hat{\boldsymbol{m}}_{ij}\}$ over $N$ videos and reconstructs leaderboard scores $\mathbf{s}={(s_1,\dots,s_N)}^\top$, where $s_i$ denotes the score of video $i$. Each compared pair $(i,j)$ contributes one linear equation
\begin{equation}
\label{eq:lsq-pair}
s_i - s_j \;\approx\; \hat{\boldsymbol{m}}_{ij}\,.
\end{equation}
Stacking these equations over the compared-pair set $\mathcal{P}$ yields an overdetermined system. Let the pairs be indexed by $k=1,\ldots,M$, where $M=|\mathcal{P}|$, and let the $k$-th pair have indices $i_k$ and $j_k$. The signed comparison matrix $\mathbf{A}\in\mathbb{R}^{M\times N}$ is
\begin{equation}
A_{k,n}=\begin{cases}
+1, & n=i_k,\\
-1, & n=j_k,\\
0, & \text{otherwise}.
\end{cases}
\end{equation}
Each row of $\mathbf{A}$ therefore encodes one relative comparison, and multiplying it by the score vector extracts the corresponding score difference. The stacked matrix form is
\begin{equation}
\label{eq:linear-system}
\mathbf{A}\mathbf{s}
=
\begin{bmatrix}
s_{i_1}-s_{j_1}\\
s_{i_2}-s_{j_2}\\
\vdots\\
s_{i_M}-s_{j_M}
\end{bmatrix}
\;\approx\;
\begin{bmatrix}
\hat{\boldsymbol{m}}_{i_1,j_1}\\
\hat{\boldsymbol{m}}_{i_2,j_2}\\
\vdots\\
\hat{\boldsymbol{m}}_{i_M,j_M}
\end{bmatrix}
=\mathbf{b}\,.
\end{equation}
Because predicted margins may be noisy and globally inconsistent, leaderboard scores are obtained by the constrained least-squares (LSQ) problem~\citep{jiang2011statistical}:
\begin{equation}
\label{eq:lsq}
\mathbf{s}^{*}
\;=\;
\operatornamewithlimits{argmin}_{\mathbf{s}\in\mathbb{R}^N}
\|\mathbf{A}\mathbf{s}-\mathbf{b}\|_2^2,
\quad\text{s.t.}\;\;
\textstyle\sum_i s_i=0\,.
\end{equation}
For a connected comparison graph, the zero-mean constraint removes the global shift ambiguity, and the solution satisfies $(\mathbf{A}^\top\!\mathbf{A})\,\mathbf{s}^{*}=\mathbf{A}^\top\mathbf{b}$ under this gauge constraint.
Thus, the leaderboard is \textbf{anchor-free}: scores are recovered from predicted-margin consistency without a reference sample. As $|\mathcal{P}|$ increases, the system averages per-pair noise and accumulates finer relative evidence.
\subsection{Supervision for Quality Differences}
\label{sec:method:supervision}

The predictive margin in Sec.~\ref{sec:method:pairwise} formulates VQA as supervised quality-difference learning, where the model explains and predicts the quality gap for each video pair. This section introduces the unified response format for training, followed by the motion-aware VLM, regression head, and supervised objective.

\paragraph{Response Format.}
\label{sec:method:supervision:format}

For every pairwise prompt $x_{ij}$, the model follows a fixed response template
\begin{equation}
\label{eq:response-format}
y_{ij}
=
\texttt{<think>}\;\mathbf{c}_{ij}\;\texttt{</think><answer><query>},
\end{equation}
where $\mathbf{c}_{ij}$ is the comparison rationale enclosed in the \texttt{<think>} block. The \texttt{<answer>} token marks the end of generated text, after which the non-decoded \texttt{<query>} placeholder is appended.

\paragraph{Motion-Aware VLM Backbone.}
\label{sec:method:supervision:motion}

Sparse visual frames provide limited inter-frame information for assessing temporal artifacts such as motion blur, frame drops, and judder.
Inspired by VQA$^{2}$~\citep{vqa2scorer}, we integrate a frozen SlowFast R50~\citep{feichtenhofer2019slowfast} encoder through an auxiliary embedding pathway.
For each video, we densely sample $T$ frames and spatially pool the encoder's fast-pathway features to obtain frame-level motion representations $\tilde{\mathbf{u}}_1,\ldots,\tilde{\mathbf{u}}_T$.
The pathway's only trainable component, a lightweight projector $g_{\theta}$, maps these representations into motion tokens $\mathbf{u}_t$ in the LLM embedding space:
\begin{equation}
\label{eq:motion-projector}
\mathbf{u}_t = g_{\theta}(\tilde{\mathbf{u}}_t), \qquad t=1,\ldots,T.
\end{equation}

\paragraph{Regression Head for Quality-Difference Prediction.}
\label{sec:method:supervision:reghead}

To avoid \emph{output quantization} from numeric token decoding~\citep{qalign}, a lightweight trainable MLP head $r_{\phi}$ maps the hidden state $\mathbf{h}_{ij}\in\mathbb{R}^{d}$ at \texttt{<query>} directly to a continuous signed margin:
\begin{equation}
\label{eq:reghead}
\hat{\boldsymbol{m}}_{ij} = r_{\phi}(\mathbf{h}_{ij})\,.
\end{equation}
Since \texttt{<query>} follows the comparison rationale, margin prediction is conditioned on this whole reasoning process.

\paragraph{Supervised Quality-Difference Learning.}
\label{sec:method:supervision:sft}

Before reinforcement learning, we jointly warm-start the language model, motion projector, and regression head on examples $(v_i, v_j, \boldsymbol{m}_{ij}^{\star}, \mathbf{c}_{ij})$ with rationales annotated by Gemini-3-Pro~\citep{gemini3pro}.
Using the template in Eq.~\eqref{eq:response-format}, the next-token prediction loss $\mathcal{L}_{\text{NTP}}$ supervises textual tokens through \texttt{<answer>}.
For a mini-batch of size $B$, the regression loss is
\begin{equation}
\label{eq:reg-loss}
\mathcal{L}_{\text{reg}}
\;=\;
\frac{1}{B}\sum_{k=1}^{B}
\left\|\hat{\boldsymbol{m}}^{(k)} - \boldsymbol{m}^{\star(k)}\right\|_2^2,
\end{equation}
Here $k$ indexes training pairs. The joint supervised objective is
\begin{equation}
\label{eq:sft-loss}
\mathcal{L}_{\text{SFT}}
\;=\;
\mathcal{L}_{\text{NTP}}
\;+\;
\lambda_{\text{reg}}\,\mathcal{L}_{\text{reg}}\,,
\end{equation}
where $\lambda_{\text{reg}}$ is a hyperparameter controlling the strength of the supervised regression term.

\subsection{Reinforced Reasoning}
\label{sec:method:reinforced}
\label{sec:method:reinforced:grpo}

The supervised objective above provides a strong initialization, but it only teaches the model to imitate fixed reasoning traces.
To let the policy explore better comparison rationales while keeping the predicted margin calibrated, we apply \textbf{Margin-Coupled GRPO (MC-GRPO)} based on Group Relative Policy Optimization~\citep{shao2024deepseekmath}.
MC-GRPO updates the rationale-generating policy using group-relative rewards, while employing an auxiliary regression gradient to ensure the \texttt{<query>} hidden state remains predictive of the continuous quality gap.
\vspace{-5pt}
\paragraph{Rollout and Reward.}
For a pairwise prompt $x_{ij}$, the rollout policy $\pi_{\boldsymbol{\theta}_{\mathrm{old}}}$ samples $G$ responses $o_1,\ldots,o_G$ in the format of Eq.~\eqref{eq:response-format} up to \texttt{<answer>}. For each $o_g$, the \texttt{<query>} placeholder is appended and the regression head predicts $\hat{\boldsymbol{m}}_{ij,g}$.
Given the MOS-derived target margin $\boldsymbol{m}_{ij}^{\star}$, the reward for the $g$-th response is
\begin{equation}
\label{eq:reward}
\begin{aligned}
R_g ={}&
\exp\!\left(-\alpha\left\|\hat{\boldsymbol{m}}_{ij,g}
-\boldsymbol{m}_{ij}^{\star}\right\|_2^2\right)\\
&+\begin{cases}
w_{\mathrm{f}}, & \text{if } o_g \text{ has a valid format},\\
0, & \text{otherwise}.
\end{cases}
\end{aligned}
\end{equation}
Here $\alpha$ controls sensitivity to squared margin error, and $w_{\mathrm{f}}$ is the format reward for Eq.~\eqref{eq:response-format}. GRPO estimates the relative advantage by normalizing rewards within the group:
\begin{equation}
\label{eq:grpo-advantage}
\hat{A}_g
=
\frac{R_g-\operatorname{mean}(\{R_1,\ldots,R_G\})}
{\operatorname{std}(\{R_1,\ldots,R_G\})+\epsilon_{\mathrm{std}}}
\,,
\end{equation}
where $\epsilon_{\mathrm{std}}$ is a small constant for numerical stability.

\paragraph{GRPO Objective.}
Let $\pi_{\boldsymbol{\theta}}$, $\pi_{\boldsymbol{\theta}_{\mathrm{old}}}$, and $\pi_{\mathrm{ref}}$ be the current, rollout, and reference policies. For each response, the importance ratio is $\rho_g=\frac{\pi_{\boldsymbol{\theta}}(o_g\mid x_{ij})}{\pi_{\boldsymbol{\theta}_{\mathrm{old}}}(o_g\mid x_{ij})}$. The clipped GRPO objective is
\begin{equation}
\label{eq:grpo-objective}
\begin{aligned}
\mathcal{J}_{\mathrm{GRPO}} ={}&
\mathbb{E}\Biggl[\frac{1}{G}\sum_{g=1}^{G}\Bigl(
\min\bigl(\rho_g\hat{A}_g,\\
&\quad\operatorname{clip}(\rho_g,1-\varepsilon,1+\varepsilon)
\hat{A}_g\bigr)\\
&\quad-\beta D_{\mathrm{KL}}(\pi_{\boldsymbol{\theta}}\|\pi_{\mathrm{ref}})
\Bigr)\Biggr].
\end{aligned}
\end{equation}
Here $\mathbb{E}[\cdot]$ denotes expectation over training prompts and sampled responses, $\varepsilon$ is the clipping threshold, $D_{\mathrm{KL}}$ denotes KL divergence, and $\beta$ controls the KL penalty. The corresponding optimization loss is $\mathcal{L}_{\mathrm{GRPO}}=-\mathcal{J}_{\mathrm{GRPO}}$.

\paragraph{Margin-coupled Update.}
The policy loss changes the generated rationale, so the final hidden state at \texttt{<query>} may drift away from the representation needed for continuous scoring. MC-GRPO therefore reuses the rollout predictions in the margin MSE of Eq.~\eqref{eq:reg-loss} and optimizes it together with the policy loss.
Let $\boldsymbol{\theta}$ and $\boldsymbol{\phi}$ denote the shared VLM/projector parameters and the regression-head parameters, respectively. A single MC-GRPO update is defined as:
\begin{equation}
\label{eq:grpo-grad}
\begin{aligned}
\boldsymbol{\theta}
&\leftarrow
\boldsymbol{\theta}
-\eta\left(
\nabla_{\boldsymbol{\theta}}\mathcal{L}_{\mathrm{GRPO}}
+\lambda_{\mathrm{aux}}
\nabla_{\boldsymbol{\theta}}\mathcal{L}_{\text{reg}}
\right),
\\
\boldsymbol{\phi}
&\leftarrow
\boldsymbol{\phi}
-\eta_{\phi}\lambda_{\mathrm{aux}}
\nabla_{\boldsymbol{\phi}}\mathcal{L}_{\text{reg}}.
\end{aligned}
\end{equation}
Here, $\nabla_{\boldsymbol{\theta}}$ and $\nabla_{\boldsymbol{\phi}}$ denote gradients with respect to the VLM parameters and regression-head parameters, respectively. The learning rates $\eta$ and $\eta_{\phi}$ apply to these two parameter sets, respectively, and $\lambda_{\mathrm{aux}}$ weights the auxiliary margin-regression loss $\mathcal{L}_{\text{reg}}$ defined in Eq.~\eqref{eq:reg-loss}. The policy loss $\mathcal{L}_{\mathrm{GRPO}}$ optimizes rationale generation, while the auxiliary regression gradients jointly refine the \texttt{<query>} representation and the regression head.

\section{Experiments}\label{sec:experiments}
\begin{table*}[!t]
  \caption{Quantitative results on in-domain and out-of-domain VQA benchmarks~\citep{lsvq,hosu2017konvid,sinno2019large,yu2022perceptual,li2019avc,gao2023vdpve}. The training data column reports the training set used by each method. The upper and lower panels report SRCC and PLCC, respectively. Higher values indicate better performance. \textbf{Bold} and \underline{underlined} values denote the best and second-best results, respectively.}%
  \label{tab:main-results}
  \centering
  \footnotesize
  \setlength{\tabcolsep}{2.0pt}
  \renewcommand{\arraystretch}{1.04}
  \resizebox{\textwidth}{!}{%
  \begin{tabular}{@{}lc*{9}{>{\centering\arraybackslash}p{1.40cm}}@{}}
    \toprule
    \multirow{2}{*}{Method} & \multirow{2}{*}{\mbox{Training data}} & \multicolumn{3}{c}{In-domain} & \multicolumn{6}{c}{Out-of-domain} \\
    \cmidrule(lr){3-5} \cmidrule(l){6-11}
    & & {\fontsize{8}{9.6}\selectfont\mbox{LSVQ-Test}} & {\fontsize{8}{9.6}\selectfont\mbox{LSVQ-1080p}} & {\fontsize{8}{9.6}\selectfont\mbox{Avg.}} & {\fontsize{8}{9.6}\selectfont\mbox{KoNViD-1K}} & {\fontsize{8}{9.6}\selectfont\mbox{LIVE-VQC}} & {\fontsize{8}{9.6}\selectfont\mbox{YT-Gaming}} & {\fontsize{8}{9.6}\selectfont\mbox{W-IVC-4K}} & {\fontsize{8}{9.6}\selectfont\mbox{VDPVE}} & {\fontsize{8}{9.6}\selectfont\mbox{Avg.}} \\
    \midrule
    \rowcolor{gray!15}
    \multicolumn{11}{@{}l}{\textbf{SRCC}} \\
    \addlinespace[1pt]
    \multicolumn{11}{@{}l}{\emph{Training-free / unsupervised quality models}} \\
    NIQE & None & 0.442 & 0.489 & 0.466 & 0.541 & 0.596 & 0.240 & 0.048 & 0.415 & 0.368 \\
    VIIDEO & None & 0.080 & 0.009 & 0.045 & 0.299 & 0.033 & 0.077 & 0.114 & 0.298 & 0.164 \\
    STEM & None & 0.206 & 0.434 & 0.320 & 0.619 & 0.594 & 0.103 & 0.184 & 0.389 & 0.378 \\
    CLIP-IQA & None & 0.438 & 0.553 & 0.496 & 0.696 & 0.704 & 0.358 & 0.133 & 0.415 & 0.461 \\
    \midrule
    \multicolumn{11}{@{}l}{\emph{Supervised VQA/IQA models}} \\
    SimpleVQA & LSVQ (28K) & 0.864 & 0.756 & 0.810 & 0.861 & 0.762 & 0.657 & 0.379 & 0.643 & 0.660 \\
    FAST-VQA & LSVQ (28K) & 0.880 & 0.781 & 0.831 & 0.859 & \textbf{0.826} & 0.631 & 0.327 & 0.611 & 0.651 \\
    DOVER & LSVQ (28K) & 0.878 & 0.782 & 0.830 & 0.874 & \underline{0.817} & 0.647 & 0.368 & 0.627 & 0.667 \\
    MinimalisticVQA & LSVQ (28K) & \underline{0.885} & 0.792 & 0.839 & 0.862 & 0.775 & 0.686 & 0.459 & 0.639 & 0.684 \\
    Q-Align & fused (287K) & \textbf{0.886} & 0.761 & 0.824 & \underline{0.876} & 0.783 & 0.611 & 0.414 & 0.639 & 0.665 \\
    VQA$^{2}$-Scorer & VQA$^{2}$-ID (157K) & 0.878 & 0.794 & 0.836 & \textbf{0.881} & 0.785 & 0.613 & 0.415 & 0.684 & 0.676 \\
    \midrule
    \multicolumn{11}{@{}l}{\emph{Reasoning or reinforcement-learning based LMM evaluators}} \\
    Q-Insight & LSVQ (28K) & 0.644 & 0.601 & 0.623 & 0.751 & 0.624 & 0.310 & 0.218 & 0.547 & 0.490 \\
    VisualQuality-R1 & LSVQ (28K) & 0.795 & 0.716 & 0.756 & 0.784 & 0.732 & 0.472 & 0.227 & 0.622 & 0.567 \\
    VQ-Insight & LSVQ \& KoNIQ (38K) & 0.875 & 0.786 & 0.831 & 0.875 & 0.790 & N/A & N/A & N/A & N/A \\
    VQAThinker & LSVQ (28K) & 0.883 & \underline{0.798} & \underline{0.841} & \textbf{0.881} & 0.808 & \underline{0.767} & \underline{0.573} & \underline{0.706} & \underline{0.747} \\
    \textbf{VersusQ (ours)} & LSVQ (28K) & 0.874 & \textbf{0.818} & \cellcolor{red!15}\textbf{0.846} & 0.875 & 0.811 & \textbf{0.803} & \textbf{0.615} & \textbf{0.719} & \cellcolor{red!15}\textbf{0.764} \\
    \midrule
    \addlinespace[1pt]
    \rowcolor{gray!15}
    \multicolumn{11}{@{}l}{\textbf{PLCC}} \\
    \addlinespace[1pt]
    \multicolumn{11}{@{}l}{\emph{Training-free / unsupervised quality models}} \\
    NIQE & None & 0.332 & 0.459 & 0.396 & 0.553 & 0.628 & 0.247 & 0.002 & 0.335 & 0.353 \\
    VIIDEO & None & 0.080 & 0.019 & 0.050 & 0.300 & 0.215 & -0.199 & 0.078 & 0.254 & 0.130 \\
    STEM & None & 0.243 & 0.381 & 0.312 & 0.627 & 0.629 & 0.111 & 0.097 & 0.313 & 0.355 \\
    CLIP-IQA & None & 0.413 & 0.505 & 0.459 & 0.651 & 0.683 & 0.384 & 0.213 & 0.446 & 0.475 \\
    \midrule
    \multicolumn{11}{@{}l}{\emph{Supervised VQA/IQA models}} \\
    SimpleVQA & LSVQ (28K) & 0.861 & 0.801 & 0.831 & 0.860 & 0.799 & 0.728 & 0.425 & 0.647 & 0.692 \\
    FAST-VQA & LSVQ (28K) & 0.880 & 0.813 & 0.847 & 0.854 & 0.845 & 0.677 & 0.363 & 0.620 & 0.672 \\
    DOVER & LSVQ (28K) & 0.866 & 0.813 & 0.840 & 0.869 & 0.840 & 0.728 & 0.418 & 0.631 & 0.697 \\
    MinimalisticVQA & LSVQ (28K) & \underline{0.882} & 0.828 & 0.855 & 0.859 & 0.821 & 0.746 & 0.502 & 0.641 & 0.714 \\
    Q-Align & fused (287K) & \textbf{0.884} & 0.822 & 0.853 & 0.878 & 0.819 & 0.681 & 0.497 & 0.649 & 0.705 \\
    VQA$^{2}$-Scorer & VQA$^{2}$-ID (157K) & 0.872 & 0.821 & 0.847 & 0.880 & 0.830 & 0.698 & 0.474 & 0.692 & 0.715 \\
    \midrule
    \multicolumn{11}{@{}l}{\emph{Reasoning or reinforcement-learning based LMM evaluators}} \\
    Q-Insight & LSVQ (28K) & 0.639 & 0.648 & 0.644 & 0.753 & 0.708 & 0.326 & 0.206 & 0.564 & 0.511 \\
    VisualQuality-R1 & LSVQ (28K) & 0.796 & 0.744 & 0.770 & 0.792 & 0.781 & 0.548 & 0.298 & 0.637 & 0.611 \\
    VQ-Insight & LSVQ \& KoNIQ (38K) & 0.876 & 0.823 & 0.850 & \underline{0.884} & 0.835 & N/A & N/A & N/A & N/A \\
    VQAThinker & LSVQ (28K) & 0.880 & \underline{0.834} & \underline{0.857} & \underline{0.884} & \underline{0.847} & \underline{0.806} & \underline{0.624} & \underline{0.716} & \underline{0.775} \\
    \textbf{VersusQ (ours)} & LSVQ (28K) & 0.878 & \textbf{0.850} & \cellcolor{red!15}\textbf{0.864} & \textbf{0.885} & \textbf{0.851} & \textbf{0.810} & \textbf{0.709} & \textbf{0.737} & \cellcolor{red!15}\textbf{0.798} \\
    \bottomrule
  \end{tabular}%
  }
\end{table*}

\begin{table}[t]
  \caption{Ablation study with mean SRCC and PLCC. In-domain averages cover LSVQ-Test and LSVQ-1080p, while out-of-domain averages cover the other five benchmarks. \textbf{Bold} denotes the best result within each group.}
  \label{tab:ablation}
  \centering
  \footnotesize
  \setlength{\tabcolsep}{2.5pt}
  \renewcommand{\arraystretch}{1.08}
  \begin{tabular*}{\columnwidth}{@{\extracolsep{\fill}}lcccc@{}}
    \toprule
    \multirow{2}{*}{Variant} & \multicolumn{2}{c}{In-domain} & \multicolumn{2}{c}{Out-of-domain} \\
    \cmidrule(lr){2-3} \cmidrule(l){4-5}
    & SRCC & PLCC & SRCC & PLCC \\
    \midrule
    \multicolumn{5}{@{}l}{\emph{Scoring target}} \\
    Pointwise Score (NTP) & 0.819 & 0.835 & 0.740 & 0.756 \\
    Pointwise Score (head) & 0.836 & 0.852 & 0.711 & 0.751 \\
    Pairwise Margin (NTP) & 0.841 & 0.860 & 0.748 & 0.784 \\
    Pairwise Margin (head) & \textbf{0.846} & \textbf{0.864} & \textbf{0.764} & \textbf{0.798} \\
    \midrule
    \multicolumn{5}{@{}l}{\emph{Score recovery}} \\
    Win-rate Ranking & 0.827 & 0.839 & 0.737 & 0.769 \\
    Elo Rating & 0.842 & 0.857 & 0.755 & 0.783 \\
    Least-squares Recovery & \textbf{0.846} & \textbf{0.864} & \textbf{0.764} & \textbf{0.798} \\
    \midrule
    \multicolumn{5}{@{}l}{\emph{Training recipe (margin + LSQ)}} \\
    SFT w/o Reasoning (18k) & 0.785 & 0.800 & 0.697 & 0.729 \\
    SFT w/ Reasoning (18k) & 0.814 & 0.839 & 0.724 & 0.760 \\
    GRPO from SFT & \textbf{0.847} & 0.863 & 0.751 & 0.788 \\
    MC-GRPO from SFT & 0.846 & \textbf{0.864} & \textbf{0.764} & \textbf{0.798} \\
    \bottomrule
  \end{tabular*}
\end{table}

\subsection{Experimental Setup}
\paragraph{Datasets and metrics.}
We evaluate VersusQ on seven public VQA benchmarks: LSVQ-test and LSVQ-1080p~\citep{lsvq}, KoNViD-1k~\citep{hosu2017konvid}, LIVE-VQC~\citep{sinno2019large}, LIVE-YT-Gaming~\citep{yu2022perceptual}, Waterloo-IVC-4K~\citep{li2019avc}, and VDPVE~\citep{gao2023vdpve}.
Unless otherwise specified, we train on the LSVQ training split and directly evaluate on all benchmarks; following VQAThinker~\citep{vqathinker}, out-of-domain datasets are tested in full without train-test splits.
We report Spearman rank-order (SRCC) and Pearson linear (PLCC) correlations with human MOS/DMOS.

\vspace{-10pt}
\paragraph{Compared methods.}
We compare against three groups of baselines.
The first group contains unsupervised or training-free quality models, including NIQE~\citep{mittal2012making}, VIIDEO~\citep{mittal2016completely}, STEM~\citep{kancharla2022completely}, and CLIP-IQA~\citep{wang2023exploring}.
The second group contains supervised VQA/IQA models, including SimpleVQA~\citep{sun2022simplevqa}, FAST-VQA~\citep{fastvqa}, DOVER~\citep{dover}, MinimalisticVQA~\citep{sun2024minimalistic}, Q-Align~\citep{qalign}, and VQA$^{2}$-Scorer~\citep{vqa2scorer}.
The third group contains recent reasoning or reinforcement-learning based LMM evaluators, including Q-Insight~\citep{qinsight}, VisualQuality-R1~\citep{visualqualityr1}, VQ-Insight~\citep{zhang2025vqinsight}, and VQAThinker~\citep{vqathinker}.
\vspace{-10pt}
\paragraph{Implementation details.}
VersusQ is initialized from Qwen3-VL-4B-Instruct~\citep{qwen3vl} and augmented with the SlowFast R50 motion encoder~\citep{feichtenhofer2019slowfast}. 
For each video in one pair, we uniformly sample 16 visual frames and extract 64 SlowFast motion tokens as input. The maximum sequence length is set to 2048.
During supervised warm-up, the LLM is fully fine-tuned for one epoch on 18k LSVQ video pairs with a global batch size of 8. We utilize the AdamW optimizer with a cosine decay scheduler. The peak learning rates are $1\times10^{-5}$ for the backbone and $1\times10^{-4}$ for both the motion projector and regression head. In Eq.~\eqref{eq:sft-loss}, $\lambda_{\text{reg}}$ is set to 0.5.
Starting from the SFT checkpoint, MC-GRPO trains for one epoch on 60k LSVQ pairwise prompts with 4 rollouts per prompt. The learning rate is $1\times10^{-6}$, with auxiliary margin weight $\lambda_{\mathrm{aux}}=0.5$. Reward weights are 1.0 for margin-regression accuracy and 0.2 for format validity. During rollouts, we set temperature to 1.0.
The visual and motion encoders are kept frozen throughout both stages.
\vspace{-10pt}
\paragraph{Training and evaluation pair.}
Training pairs are randomly sampled from the LSVQ training split and labeled with signed MOS differences (Eq.~\eqref{eq:margin}).
For evaluation on a benchmark with $N$ videos, we use $5N$ comparisons for stable rank recovery, divided into 50 batches of $N/10$ comparisons each.
The first batch uses randomly balanced pairing.
Later batches prioritize low-degree videos and then select partners with nearby provisional scores, which are reconstructed solely from model-predicted margins using Eq.~\eqref{eq:lsq}.
The final leaderboard is jointly recovered from all accumulated comparisons on the connected graph.
Ground-truth MOS/DMOS is used only to compute metrics.
Section~\ref{sec:comparison-budget} analyzes the comparison budget.

\subsection{Main Results}
Table~\ref{tab:main-results} summarizes the benchmark comparison.
VersusQ achieves the highest mean SRCC and PLCC on both in-domain and out-of-domain benchmarks.
On the two in-domain LSVQ benchmarks, VersusQ achieves mean SRCC of 0.846 and PLCC of 0.864, compared with the second-best results of 0.841 and 0.857, respectively.

Without target-dataset fine-tuning, VersusQ achieves mean out-of-domain SRCC of 0.764 and PLCC of 0.798, exceeding the second-best results of 0.747 and 0.775 by \textbf{0.017} and \textbf{0.023}, respectively.
It achieves the best PLCC on all five out-of-domain datasets and the best SRCC on YT-Gaming, W-IVC-4K, and VDPVE.
The largest PLCC gain over the second-best result occurs on W-IVC-4K: 0.709 versus 0.624, an absolute increase of \textbf{0.085} in the rounded results.
SRCC on this dataset reaches 0.615, exceeding the second-best result of 0.573 by \textbf{0.042}.
On YT-Gaming, SRCC reaches 0.803, an absolute gain of \textbf{0.036} over the second-best result of 0.767.
On KoNViD and LIVE-VQC, VersusQ achieves the highest PLCC of 0.885 and 0.851, respectively.
Together, these results demonstrate strong out-of-domain generalization, highlighting the generalization advantages of our pairwise quality assessment approach.

\begin{figure}[t]
  \centering
  \includegraphics[width=\linewidth]{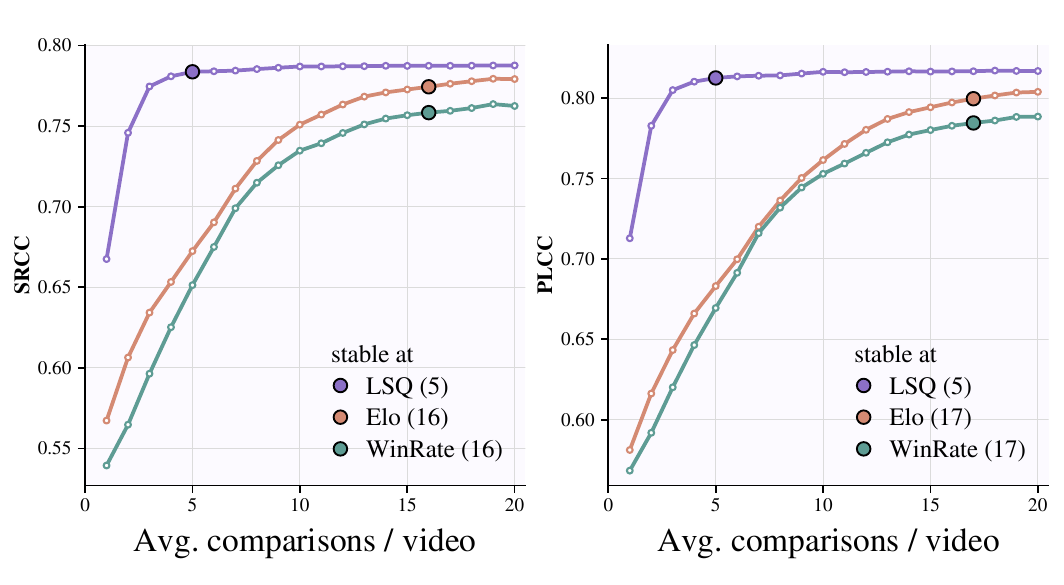}
  \caption{\textbf{Comparison budget and leaderboard convergence.} Mean SRCC (left) and PLCC (right) over seven benchmarks as the average number of comparisons per video increases. Filled circles indicate convergence within 0.005 of each metric's final value.}
  \label{fig:mean-pair-convergence}
\end{figure}

\subsection{Ablation Studies}
Table~\ref{tab:ablation} compares scoring targets, score recovery, and training recipes using mean SRCC and PLCC across in-domain and out-of-domain benchmarks.

\vspace{-10pt}
\paragraph{Scoring target.}
Pairwise NTP improves over pointwise NTP in all four domain averages.
The margin regression head further improves mean SRCC to 0.846 in-domain and 0.764 out-of-domain, with corresponding mean PLCC values of 0.864 and 0.798.
The cross-domain gains suggest that relative quality differences provide transferable supervision targets. Direct continuous prediction with the regression head further improves fine-grained quality assessment.
\vspace{-20pt}
\paragraph{Score recovery from margins.}
On the same comparisons, LSQ achieves the highest mean SRCC and PLCC in both domains.
For out-of-domain evaluation, LSQ achieves mean SRCC of 0.764 and PLCC of 0.798. Elo~\citep{elo1978rating} and win-rate ranking~\citep{bradley1952rank} reach SRCC of 0.755 and 0.737, and PLCC of 0.783 and 0.769, respectively. 
This advantage highlights the value of retaining quality-gap magnitudes alongside preference directions during global score reconstruction, since the predicted margin contributes additional information entropy for this quality difference.
\vspace{-10pt}
\paragraph{Training recipe.}
Reasoning-supervised SFT improves all four domain averages over SFT without reasoning, and both GRPO recipes improve further without requiring any additional costly reasoning-process annotations.
MC-GRPO yields the best mean out-of-domain SRCC (0.764) and PLCC (0.798), as well as the best mean in-domain PLCC (0.864), but slightly lower in-domain SRCC than standard GRPO (0.846 versus 0.847).
Together, these findings favor reasoning-guided warm-up followed by joint policy and margin optimization for cross-domain generalization.
\vspace{-10pt}
\paragraph{Anchor-based vs. anchor-free inference.}
Table~\ref{tab:anchor-based} compares anchor-based variants with $K=1,4,8$ against our anchor-free method.
Even $K=1$ achieves competitive performance with one model call per video, matching the call count of pointwise inference.
Additional anchors mainly improve out-of-domain performance, with diminishing gains beyond $K=4$.
Anchor-free inference achieves the best correlations, with its largest advantage in out-of-domain PLCC.
This may reflect reduced dependence on fixed references and joint recovery from multiple pairwise constraints, which can mitigate local comparison errors.
\begin{table}[t]
  \caption{Anchor-based and anchor-free score recovery with mean SRCC and PLCC. In-domain averages cover LSVQ-Test and LSVQ-1080p; out-of-domain averages cover the other five benchmarks. Anchor-free (ours) is the final-model reference. \textbf{Bold} denotes the best result.}
  \label{tab:anchor-based}
  \centering
  \footnotesize
  \setlength{\tabcolsep}{2.5pt}
  \renewcommand{\arraystretch}{1.08}
  \begin{tabular*}{\columnwidth}{@{\extracolsep{\fill}}lcccc@{}}
    \toprule
    \multirow{2}{*}{Variant} & \multicolumn{2}{c}{In-domain} & \multicolumn{2}{c}{Out-of-domain} \\
    \cmidrule(lr){2-3} \cmidrule(l){4-5}
    & SRCC & PLCC & SRCC & PLCC \\
    \midrule
    Anchor $K=1$ & 0.8442 & 0.8545 & 0.7525 & 0.7640 \\
    Anchor $K=4$ & 0.8424 & 0.8598 & 0.7608 & 0.7773 \\
    Anchor $K=8$ & 0.8421 & 0.8597 & 0.7622 & 0.7778 \\
    \midrule
    Anchor-free (ours) & \textbf{0.8459} & \textbf{0.8642} & \textbf{0.7645} & \textbf{0.7982} \\
    \bottomrule
  \end{tabular*}
\end{table}

\begin{figure*}[!t]
  \centering
  \includegraphics[width=\textwidth]{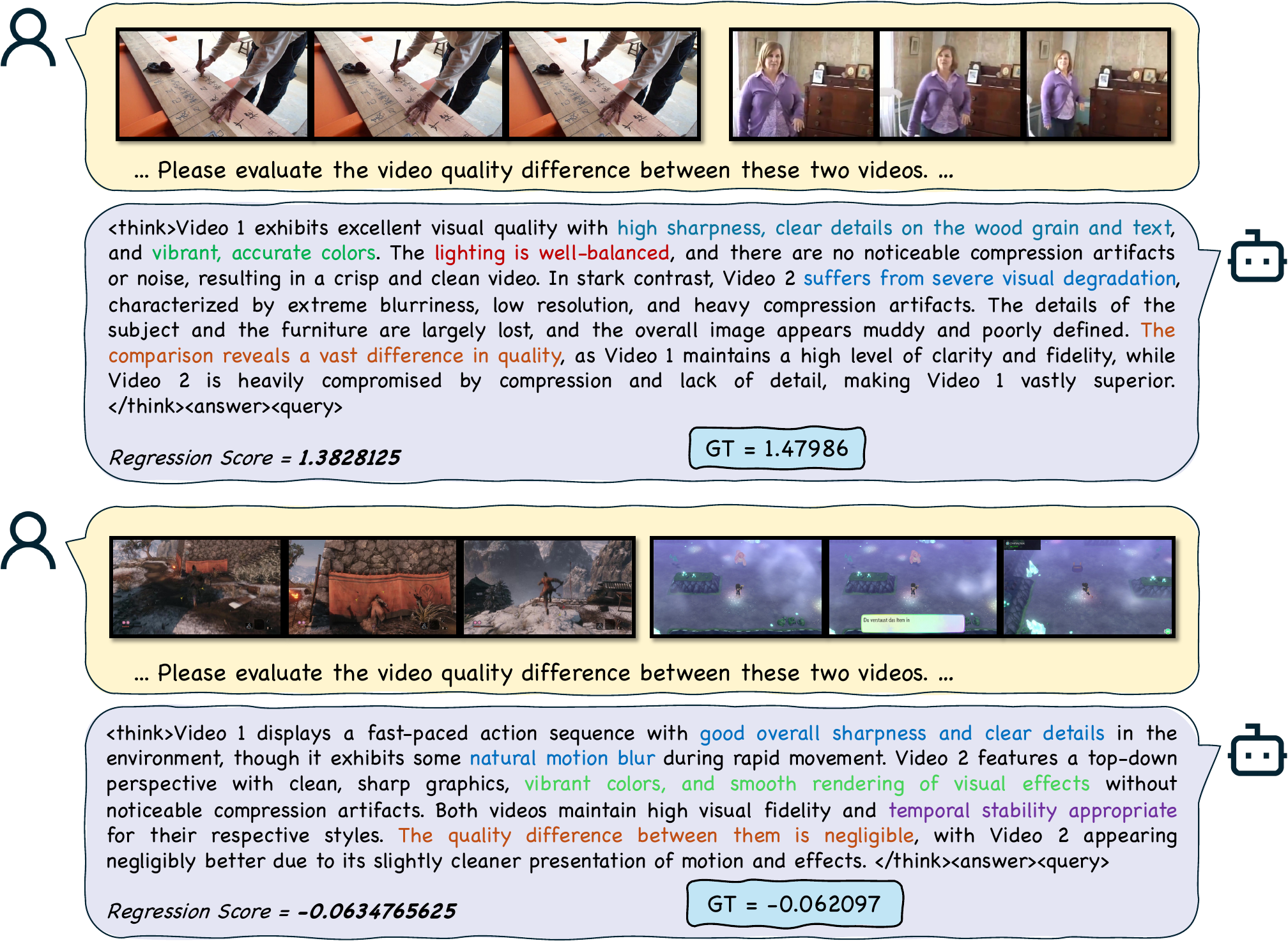}
  \caption{\textbf{Qualitative case study of VersusQ on video pairs.} The model contrasts visual evidence from each pair and predicts a continuous margin close to the ground-truth quality difference. The two cases show a large quality gap (top) and near-equal quality (bottom).}
  \label{fig:qualitative-case}
\end{figure*}

\raggedbottom
\subsection{Comparison Budget and Convergence}\label{sec:comparison-budget}

Beyond final accuracy, we examine how quickly the score recovery methods converge as the average number of comparisons per video increases.
Figure~\ref{fig:mean-pair-convergence} reports mean SRCC and PLCC over seven benchmarks as progressively more queried pairs are used for score recovery.
LSQ reaches stable SRCC and PLCC at an average of 5 comparisons per video, while Elo and WinRate require 16 for SRCC and 17 for PLCC. The observed convergence at an average of 5 comparisons per video supports the evaluation budget adopted in the evaluation protocol.
These results also highlight the benefit of our margin prediction paradigm, which captures both the direction and magnitude of quality differences and enables faster convergence with fewer comparisons when combined with LSQ score recovery.
Dataset-wise curves are provided in the supplementary material.

\subsection{Qualitative Analysis}
Figure~\ref{fig:qualitative-case} presents representative comparisons from the evaluation set, featuring an in-domain case on the top and an out-of-domain case on the bottom. In both scenarios, the generated rationale highlights specific visual differences between the two videos, such as sharpness, color fidelity, lighting, motion, and compression artifacts, while the regression head outputs a continuous margin that closely aligns with the ground-truth score difference.

\FloatBarrier
\section{Conclusion}
\label{sec:conclusion}
This paper presented VersusQ, a pairwise margin reasoning framework for generalizable video quality assessment with large multimodal models. VersusQ shifts VQA from pointwise absolute score fitting to direct video comparison, where the model reasons about visual and temporal quality differences and predicts a signed continuous margin. These local margins are then consolidated into globally comparable quality scores through anchor-free least-squares recovery. To bridge interpretable relational rationales and fine-grained numerical differences, VersusQ further introduces MC-GRPO, which jointly optimizes rollout-based reasoning and continuous margin regression. Experiments across seven public VQA benchmarks show that this design achieves state-of-the-art performance, improves cross-domain generalization, and provides reliable fine-grained ranking under heterogeneous evaluation scenarios.

\vspace{5pt}
\noindent\textbf{Limitations.}
VersusQ requires multiple pairwise evaluations to construct a leaderboard, which may limit its use in latency-sensitive settings. Pairwise differencing cancels shared additive rating offsets, but MOS-derived margins may still retain rating-scale differences and annotation noise. Future work will explore more efficient pair selection and direct human pairwise supervision.

{
    \small
    \bibliographystyle{ieeenat_fullname_etal}
    \bibliography{main}
}

\clearpage
\maketitlesupplementary
\appendix
\setcounter{figure}{0}
\setcounter{table}{0}
\renewcommand{\thefigure}{S\arabic{figure}}
\renewcommand{\thetable}{S\arabic{table}}

This supplementary material provides details omitted from the main paper for space.
Sec.~\ref{sec:appendix_cot_prompt} describes the rationale annotation used for supervised warm-up.
Sec.~\ref{sec:appendix_resources_assets} lists the full training configuration, and Sec.~\ref{sec:appendix_ablation_protocol} specifies the ablation protocols.
Sec.~\ref{sec:appendix_sparse_convergence} reports dataset-wise comparison-budget curves, and Sec.~\ref{sec:appendix_qualitative} presents additional qualitative examples.

\section{CoT Annotation}
\label{sec:appendix_cot_prompt}

Gemini-3-Pro\footnote{Google DeepMind, \href{https://storage.googleapis.com/deepmind-media/gemini/gemini_3_pro_model_evaluation.pdf}{\emph{Gemini 3 Pro: Model Evaluation -- Approach, Methodology \& Results}}, 2025.} annotates comparison rationales for the 18k LSVQ training pairs used in supervised warm-up (main paper, Sec.~\ref{sec:method:supervision:sft}). We will release this set upon acceptance.

\paragraph{Prompt-field construction.}
Each video is shown as a $3\times 3$ grid of nine uniformly sampled frames, and each pair is labeled with its ground-truth margin $d=\texttt{video1\_score}-\texttt{video2\_score}$ (Eq.~\eqref{eq:margin} of the main paper).
We set \texttt{direction\_desc} to ``Video 1 is better'' for $d>0$, ``Video 2 is better'' for $d<0$, and ``the two videos are equal in quality'' for $d=0$, and format \texttt{score\_diff\_str} as a signed value with two decimals.
The remaining fields follow the margin bins in Table~\ref{tab:appendix-prompt-bins}, using required and forbidden degree words to keep each rationale aligned with the quality gap.
Fig.~\ref{fig:appendix-cot-prompt} gives the full template.

\begin{figure*}[p]
  \centering
  \small
  \setlength{\tabcolsep}{4pt}
  \renewcommand{\arraystretch}{1.12}
  \begin{tabular}{@{}p{0.17\textwidth}p{0.21\textwidth}p{0.16\textwidth}p{0.39\textwidth}@{}}
    \toprule
    Margin magnitude & \texttt{magnitude\_desc} & \texttt{degree\_word} & Forbidden words \\
    \midrule
    $|d|<0.1$ & very similar & negligible & slight, moderate, noticeable, significant, vast \\
    $0.1\leq|d|<0.3$ & slightly different & slightly & negligible, moderate, noticeable, significant, vast \\
    $0.3\leq|d|<0.6$ & moderately different & moderately & slight, marginal, a bit, negligible, vast \\
    $0.6\leq|d|<1.0$ & significantly different & significantly & slight, moderate, negligible, marginal \\
    $|d|\geq1.0$ & very different & vastly & slight, moderate, somewhat, marginal \\
    \bottomrule
  \end{tabular}
  \captionsetup{hypcap=false}\captionof{table}{\textbf{Margin-dependent prompt fields.} Each training pair is binned by its absolute ground-truth margin $|d|$; the bin determines the magnitude description, the required degree word, and the words the rationale must avoid.}
  \label{tab:appendix-prompt-bins}
  \vspace{8pt}
\begin{annotationprompt}
You are an expert video quality assessor. You are given two videos (each represented as a 3x3 grid of sampled frames showing temporal progression) and asked to evaluate their overall visual quality difference.\par\vspace{1pt}
Image 1: Video 1's 3x3 frame grid (9 frames sampled uniformly across time)\\
Image 2: Video 2's 3x3 frame grid (9 frames sampled uniformly across time)\par\vspace{1pt}
The scoring system: Each video has a quality score ranging from 1 to 5. The score difference is computed as (Video 1 score minus Video 2 score), ranging from -4 to 4.\par\vspace{2pt}
\begin{list}{\textbullet}{\setlength{\leftmargin}{1em}\setlength{\labelsep}{0.35em}\setlength{\itemsep}{0pt}\setlength{\parsep}{0pt}\setlength{\topsep}{1pt}}
\item Positive value $\rightarrow$ Video 1 is better
\item Negative value $\rightarrow$ Video 2 is better
\item Value close to 0 $\rightarrow$ similar quality
\end{list}
The ground truth score difference is **\mbox{\texttt{\{score\_diff\_str\}}}**, meaning the two videos are \mbox{\texttt{\{magnitude\_desc\}}} and \mbox{\texttt{\{direction\_desc\}}}.\par\vspace{1pt}
The magnitude scale for reference (score difference on a 1-5 scale):\par\vspace{2pt}
\begin{list}{\textbullet}{\setlength{\leftmargin}{1em}\setlength{\labelsep}{0.35em}\setlength{\itemsep}{0pt}\setlength{\parsep}{0pt}\setlength{\topsep}{1pt}}
\item |diff| < 0.1 $\rightarrow$ negligible difference (nearly identical)
\item 0.1 $\leq$ |diff| < 0.3 $\rightarrow$ slight difference (small gap)
\item 0.3 $\leq$ |diff| < 0.6 $\rightarrow$ moderate difference (clear, meaningful gap)
\item 0.6 $\leq$ |diff| < 1.0 $\rightarrow$ significant difference (large, obvious gap)
\item |diff| $\geq$ 1.0 $\rightarrow$ vast difference (huge quality gap)
\end{list}
Please analyze both videos and generate a detailed reasoning that justifies the score difference of \mbox{\texttt{\{score\_diff\_str\}}}.\par\vspace{1pt}
Your response MUST be in valid JSON format, following this EXACT structure (no extra text or markdown formatting outside the JSON block):\par\vspace{1pt}
\begin{lstlisting}[style=promptjson]
{
  "score_difference": "{score_diff_str}",
  "reasoning": "[Your detailed analysis here. Include: 1. Description of Video 1's key visual quality characteristics (e.g., sharpness, noise, compression artifacts, color accuracy, temporal stability, exposure, motion blur) 2. Description of Video 2's key visual quality characteristics 3. Comparative analysis explaining why the two videos are {magnitude_desc} and {direction_desc}]"
}
\end{lstlisting}
Important rules:\par\vspace{2pt}
\begin{list}{\textbullet}{\setlength{\leftmargin}{1em}\setlength{\labelsep}{0.35em}\setlength{\itemsep}{0pt}\setlength{\parsep}{0pt}\setlength{\topsep}{1pt}}
\item The reasoning must be genuine and based on what you observe in the frame grids
\item Keep the reasoning concise but informative
\item Focus on overall video quality aspects: sharpness, noise/grain, compression artifacts, color fidelity, exposure, temporal consistency, motion blur, etc.
\item The degree/magnitude words in your reasoning MUST match the actual score difference. Use "\mbox{\texttt{\{degree\_word\}}}" to describe the difference. \mbox{\texttt{\{forbidden\_words\}}}
\item Ignore any subtitles, captions, watermarks, or overlaid text present in the video frames, focus only on the visual content itself
\item Write in English
\end{list}
\end{annotationprompt}
\captionof{figure}{\textbf{Pairwise CoT annotation prompt.} Fields in braces are filled per pair according to Table~\ref{tab:appendix-prompt-bins}.}
\label{fig:appendix-cot-prompt}
\end{figure*}

\section{Training Configuration}
\label{sec:appendix_resources_assets}

All experiments use NVIDIA A800 GPUs (80GB).
Supervised warm-up uses one 8-GPU node.
MC-GRPO uses four nodes: one serves asynchronous vLLM rollouts and three perform optimization.
The full two-stage pipeline takes about nine hours.
During warm-up, the signed MOS margins (Eq.~\eqref{eq:margin} of the main paper) are fed directly to the regression term of the hybrid objective in Eq.~\eqref{eq:sft-loss}.
Table~\ref{tab:appendix-training} collects the settings of both stages; the visual and motion encoders stay frozen throughout.

\begin{table}[t]
  \centering
  \footnotesize
  \setlength{\tabcolsep}{3pt}
  \renewcommand{\arraystretch}{1.08}
  \begin{tabular*}{\columnwidth}{@{\extracolsep{\fill}}lcc@{}}
    \toprule
    Setting & SFT warm-up & MC-GRPO \\
    \midrule
    \multicolumn{3}{@{}l}{\emph{Model and input}} \\
    Backbone & \multicolumn{2}{c}{Qwen3-VL-4B-Instruct} \\
    Motion encoder & \multicolumn{2}{c}{SlowFast R50 (frozen)} \\
    Frames / motion tokens per video & \multicolumn{2}{c}{16 / 64} \\
    Max sequence length & \multicolumn{2}{c}{2048} \\
    \midrule
    \multicolumn{3}{@{}l}{\emph{Optimization}} \\
    Training data & 18k pairs & 60k prompts \\
    Epochs & 1 & 1 \\
    Global batch size & 8 pairs & 48 prompts \\
    Peak learning rate & $10^{-5}$ / $10^{-4}$\,$^{\dagger}$ & $10^{-6}$ \\
    Regression weight & $\lambda_{\text{reg}}=0.5$ & $\lambda_{\text{aux}}=0.5$ \\
    \midrule
    \multicolumn{3}{@{}l}{\emph{Rollout and reward}} \\
    Rollouts per prompt $G$ & --- & 4 \\
    Sampling temperature & --- & 1.0 \\
    Margin / format reward weight & --- & 1.0 / 0.2 \\
    KL coefficient $\beta$ & --- & 0 \\
    \midrule
    \multicolumn{3}{@{}l}{\emph{Compute}} \\
    GPU nodes / total GPUs & 1 / 8 & 4 / 32 \\
    Training time & $\sim$1 hour & $\sim$8 hours \\
    \bottomrule
  \end{tabular*}
  \caption{\textbf{Training configuration of the two stages.} Shared settings span both columns; ``---'' marks settings that do not apply to a stage. $^{\dagger}$LLM backbone / motion projector and regression head, with AdamW and cosine decay.}
  \label{tab:appendix-training}
\end{table}

\section{Ablation Protocols}
\label{sec:appendix_ablation_protocol}

This section details the variants in Table~\ref{tab:ablation} of the main paper. Unless stated otherwise, all variants follow Table~\ref{tab:appendix-training} and the main evaluation protocol; in-domain and out-of-domain means average the two LSVQ splits and five out-of-domain benchmarks, respectively.

\paragraph{Scoring targets.}
Pointwise variants score one video per prompt, whereas pairwise variants predict the signed margin (Eq.~\eqref{eq:margin} of the main paper) and recover scores via LSQ. For both, NTP decodes the score as tokens, while the head variant predicts a continuous scalar from \texttt{<query>} via the MLP head (Sec.~\ref{sec:method:supervision:reghead}), using a $\sim$1:4 warm-up/GRPO ratio.

\paragraph{Score recovery.}
LSQ, Elo, and WinRate aggregate the same $5N$ comparisons and predicted margins. LSQ solves the zero-mean least-squares system (Eq.~\eqref{eq:lsq}) using continuous margins, whereas Elo and WinRate discretize outcomes into win/loss/draw with $|\hat{\boldsymbol{m}}_{ij}|<0.2$ as draws.

\paragraph{Training recipes.}
All four recipes use the pairwise margin head with LSQ recovery. The two SFT variants share the 18k pairs and settings, differing only in \texttt{<think>} rationale supervision. From reasoning-supervised SFT, GRPO-from-SFT freezes the regression head during GRPO, whereas MC-GRPO jointly updates backbone and head via Eq.~\eqref{eq:grpo-grad} of the main paper.

\twocolumn[{%
  \centering
  \includegraphics[width=0.94\textwidth]{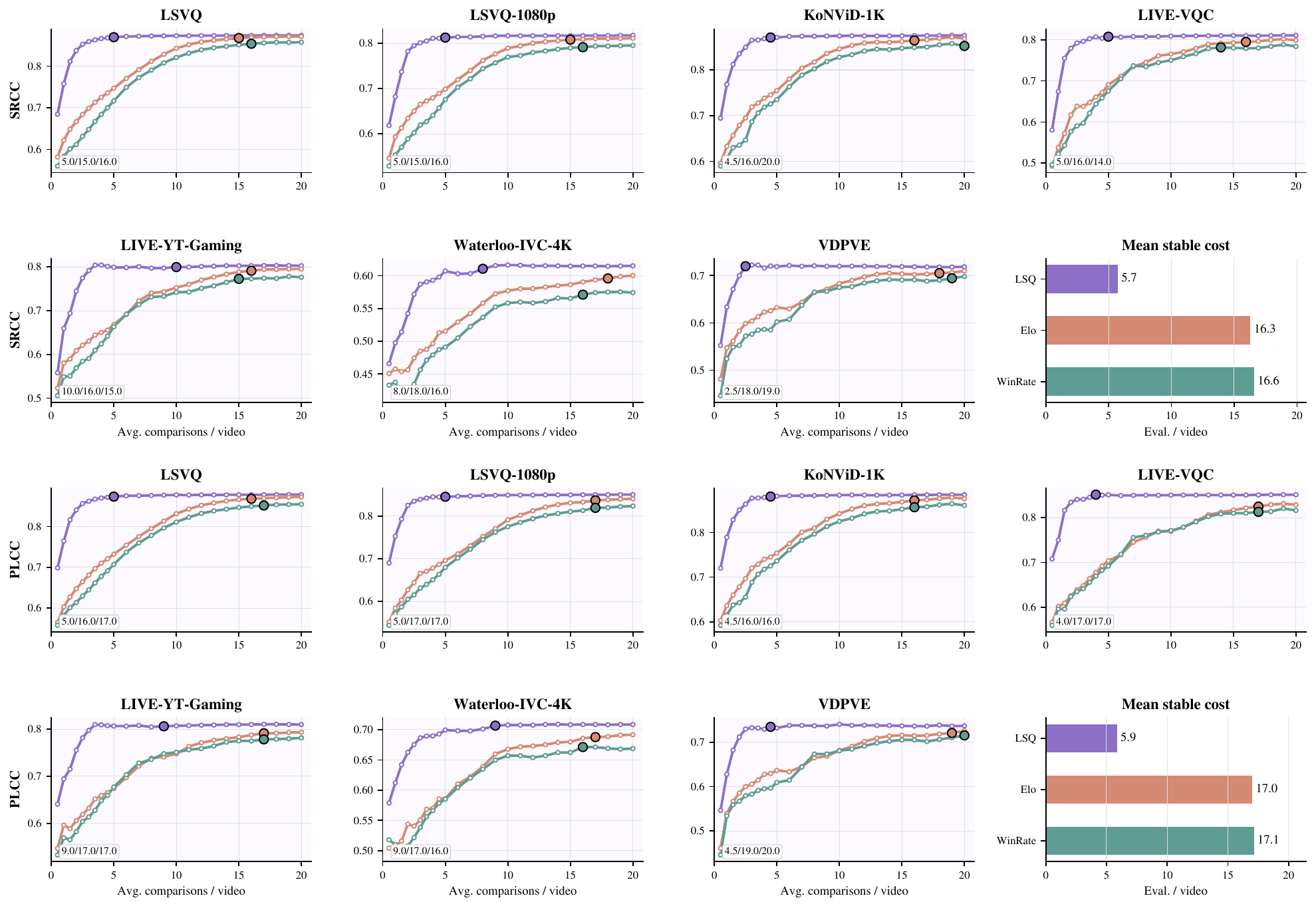}
  \captionsetup{hypcap=false}\captionof{figure}{\textbf{Dataset-wise sparse-graph convergence.} The upper and lower halves show SRCC and PLCC. Purple, orange, and teal denote LSQ, Elo, and WinRate. Filled markers indicate stability within $0.005$ of each method's final correlation; the triplet in each panel gives the stable budgets of LSQ/Elo/WinRate. Bar charts average the dataset-specific stable budgets.}
  \label{fig:appendix-method-convergence}
  \vspace{1em}
}]

\section{Dataset-wise Sparse-Graph Convergence}
\label{sec:appendix_sparse_convergence}

Fig.~\ref{fig:appendix-method-convergence} complements the mean curves in Fig.~\ref{fig:mean-pair-convergence} of the main paper (Sec.~\ref{sec:comparison-budget}) with per-dataset convergence.

\paragraph{Evaluation protocol.}
LSQ, Elo, and WinRate recover scores from growing prefixes of the same comparison stream, and we compute SRCC and PLCC against human ratings at each prefix.
The budget is $M/N$, the average number of queried pairs per video; the default $5N$ pairs give $M/N=5$ and an average graph degree of ten.
A method is stable once its correlation stays within $0.005$ of its final value.

\paragraph{Results.}
LSQ stabilizes earlier than Elo and WinRate on all seven benchmarks for both metrics.
Its mean stable budgets are 5.7 and 5.9 comparisons per video for SRCC and PLCC, versus 16.3 and 17.0 for Elo and 16.6 and 17.1 for WinRate, about 65\% fewer comparisons.
LIVE-YT-Gaming is the hardest case for LSQ, requiring 10 and 9 comparisons per video for SRCC and PLCC.
The bars average dataset-specific stable budgets, whereas the main-paper curves average correlations before thresholding, so the two summaries differ slightly.
LSQ retains the continuous signed margins in Eq.~\eqref{eq:lsq} of the main paper, which carry more information than win/draw/loss aggregation.

\section{Additional Qualitative Examples}
\label{sec:appendix_qualitative}

\begin{figure*}[p]
  \centering
  \includegraphics[width=\textwidth]{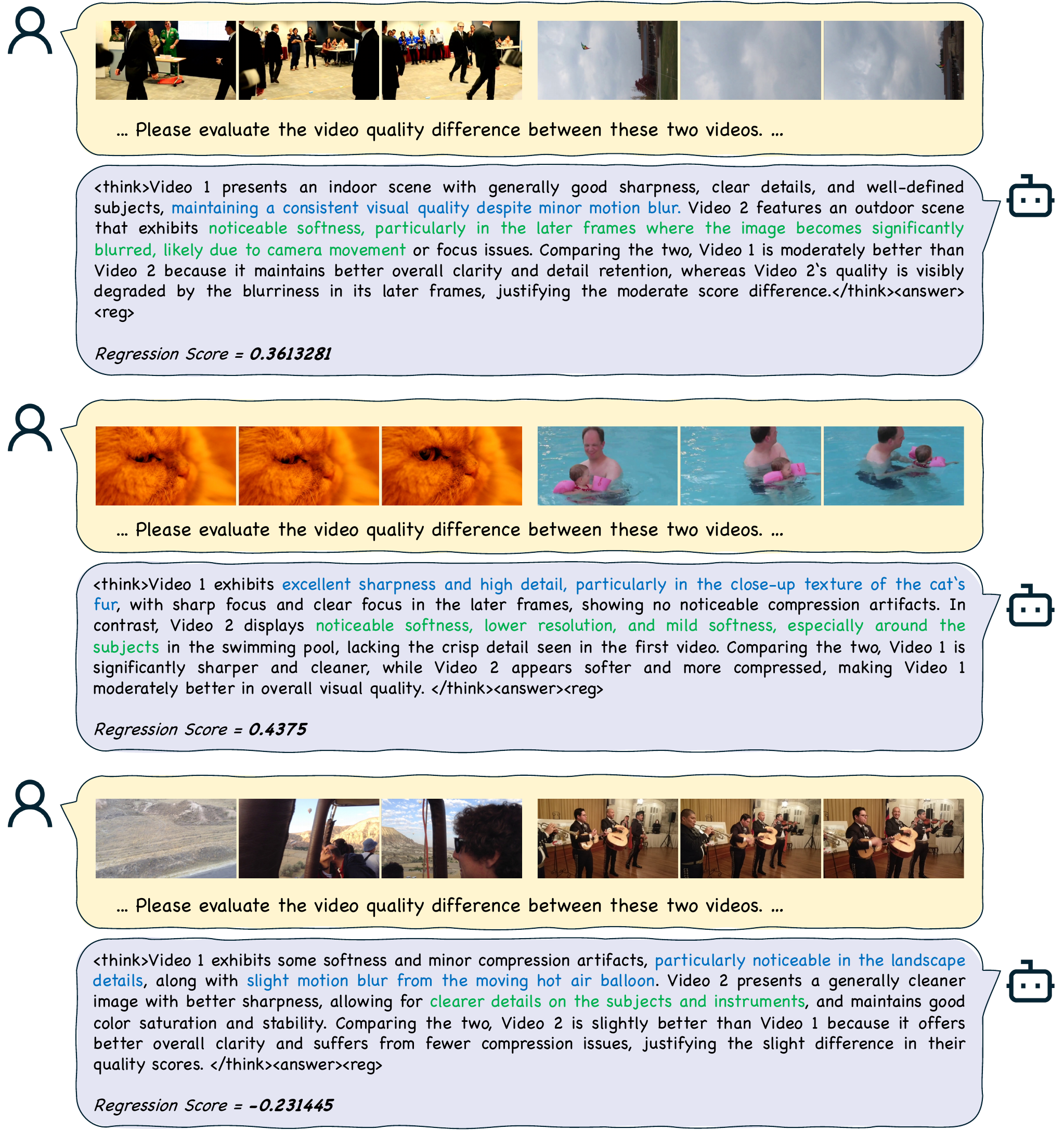}
  \caption{\textbf{Additional qualitative examples.} Comparison-oriented rationales identify perceptual differences, followed by continuous regression margins aligned with the ground-truth quality gaps.}
  \label{fig:appendix-qualitative-case}
\end{figure*}

Figs.~\ref{fig:appendix-qualitative-case} and~\ref{fig:appendix-qualitative-case-2} extend the case study in Fig.~\ref{fig:qualitative-case} of the main paper.
Each rationale separates the perceptual evidence for the two videos and explains their quality gap; the regression head then predicts a signed margin.
This makes each ranking decision auditable while retaining a continuous output for leaderboard reconstruction.

\twocolumn[{%
  \centering
  \includegraphics[width=\textwidth]{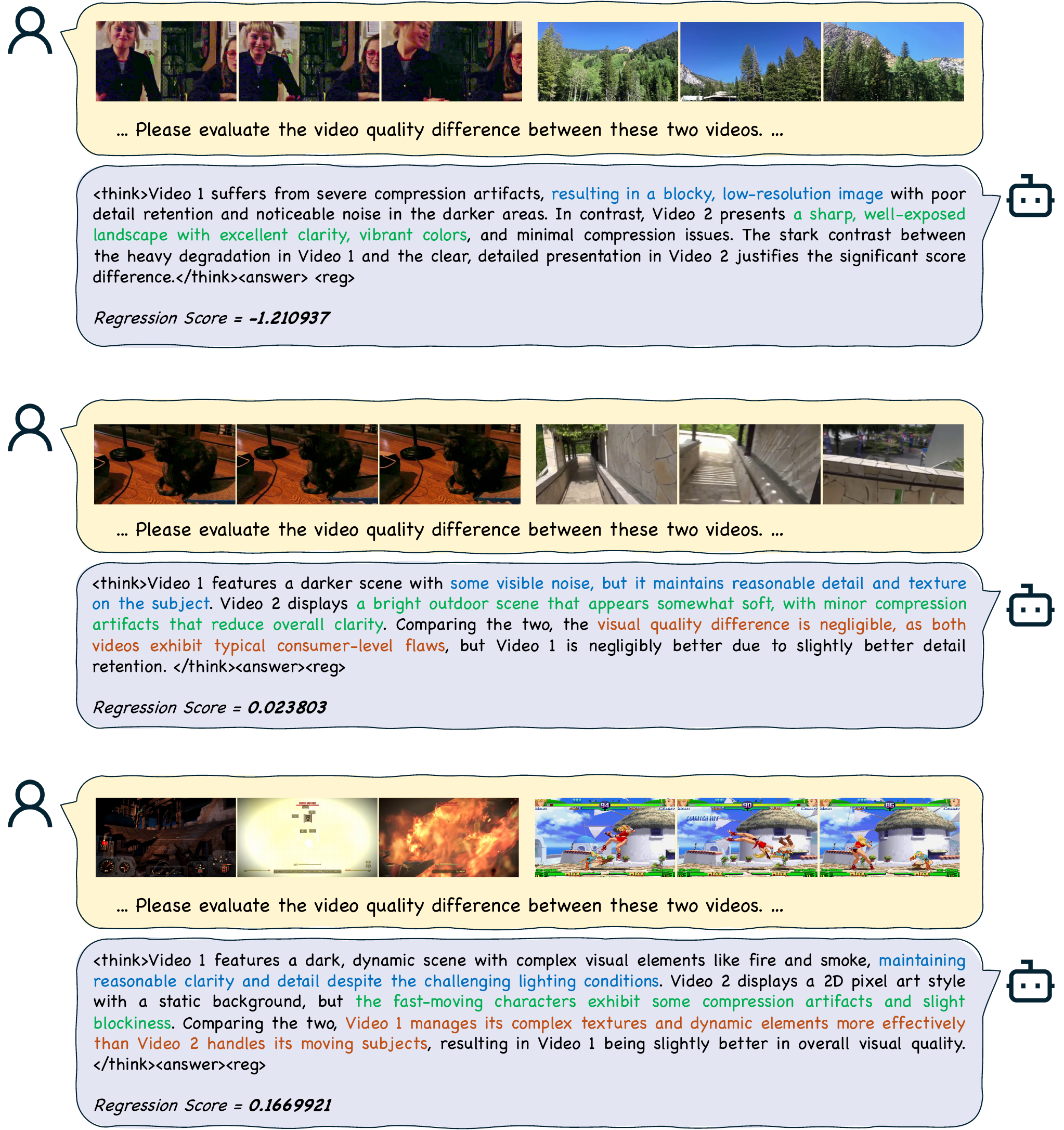}
  \captionsetup{hypcap=false}\captionof{figure}{\textbf{Additional qualitative examples (continued).} Three further pairs illustrate quality-gap reasoning and continuous signed-margin prediction.}
  \label{fig:appendix-qualitative-case-2}
  \vspace{1em}
}]

\end{document}